\documentclass[english]{edm_article}

\usepackage{amsmath}
\usepackage{booktabs}
\usepackage{array}
\usepackage{graphicx}
\usepackage{multirow}
\usepackage{subcaption}
\usepackage{algorithm}
\usepackage{algorithmic}
\usepackage{listings}
\usepackage{accsupp}
\usepackage{makecell}
\usepackage{paralist}
\usepackage{float}
\usepackage{caption}

\usepackage[a-1b]{pdfx}

\RequirePackage{natbib}

\hypersetup{
    pdftitle={Personality-Enhanced Social Recommendations in SAMI: Exploring the Role of Personality Detection in Matchmaking},     
    pdfauthor={Anon},        
    pdfsubject={Blanked}, 
    pdfkeywords={AI in Education, Social Recommendation Systems, Personality Detection in Learning Environments, AI Theory of Mind in Education, Large Language Models (LLMs) in Learning}, 
    pdfproducer={LaTeX},                  
    pdfcreator={pdflatex},                 
    colorlinks=true,                        
    linkcolor=blue,                         
    citecolor=blue,                         
    filecolor=magenta,                      
    urlcolor=cyan                           
}

\title{Personality-Enhanced Social Recommendations in SAMI: Exploring the Role of Personality Detection in Matchmaking}

\numberofauthors{4} 
\author{
 \alignauthor
 Brittany Harbison\\
        \affaddr{Georgia Institute of Technology}\\
        \affaddr{Atlanta, Georgia, USA}\\
        \email{bharbison3@gatech.edu}
 \alignauthor
 Samuel Taubman\\
        \affaddr{Georgia Institute of Technology}\\
        \affaddr{Atlanta, Georgia, USA}\\
        \email{staubman6@gatech.edu}
 \alignauthor Travis Taylor\\
        \affaddr{Georgia Institute of Technology}\\
        \affaddr{Atlanta, Georgia, USA}\\
        \email{ttaylor99@gatech.edu}
\and  
 \alignauthor Ashok K. Goel\\
        \affaddr{Georgia Institute of Technology}\\
        \affaddr{Atlanta, Georgia, USA}\\
        \email{ashok.goel@cc.gatech.edu}
}

\begin{document}
\maketitle
\begin{abstract}
Social belonging is a vital part of learning, yet online course environments present barriers to the organic formation of social groups. SAMI (Social Agent Mediated Interactions) offers one solution by facilitating student connections, but its effectiveness may be constrained by an incomplete Theory of Mind, limiting its ability to create an effective ’mental model’ of a student. One facet of this is its inability to intuit personality, which may influence the relevance of its recommendations.

To explore this gap, we examine the viability of automated personality inference by proposing a personality detection model utilizing GPT’s zeroshot capability to infer Big-Five personality traits from forum introduction posts, often encouraged in online courses. We benchmark its performance against established models, finding that while GPT models show promising results on this specific dataset, performance varies significantly across traits. We identify potential biases toward optimistic trait inference, particularly for traits with skewed distributions.

We demonstrate a proof-of-concept integration of personality detection into SAMI's entity-based matchmaking system, focusing on three traits with established connections to positive social formation: Extroversion, Agreeableness, and Openness. This work represents an initial exploration of personality-informed social recommendations in educational settings. While our implementation shows technical feasibility, significant questions remain. We discuss these limitations and outline directions for future work, examining what LLMs specifically capture when performing personality inference and whether personality-based matching meaningfully improves student connections in practice.
\end{abstract}
\section{KEYWORDS}
Educational technology, AI in online education, LLMs, personality inference, social recommendation

\section{Introduction}
Sense of community is vital for student engagement and learning, yet online courses often lead to isolation due to limited organic socialization \citep{peacock_promoting_2019,mcmillan_sense_1986}. While measures like collaborative projects and introductory posts encourage peer interaction, significant community gaps persist \citep{oyarzun_systematic_2023,vrieling-teunter_facilitating_2022}. AI-driven social recommendation systems, such as SAMI, offer a potential solution by analyzing student introduction posts to suggest connections based on shared attributes \citep{kakar2024sami}. However, systems like SAMI often lack a robust Theory of Mind (ToM), a deeper understanding of users' psychological traits, which may hinder meaningful matchmaking \citep{lex_psychology-informed_2021}.

To address this, we benchmark multiple models at the task of extracting personality traits from student text-based data and integrate this implicit personality detection model into SAMI’s matchmaking system. By leveraging trait homophily, we aim to enhance social recommendations with a more psychologically informed approach, improving the likelihood of successful peer connections. Through this work, we explore advancement in AI-supported education by investigating the role of personality-informed matchmaking in strengthening online learning communities.

\section{Background}
\subsection{Personality in Social Grouping Behavior}
The theory of homophily, or "like attracts like", is well-established in psychology and applies to friendship formation and group cohesion \citep{mcpherson_birds_2001}. While personality's role in social formation is less explored, studies have identified multiple trait-specific patterns:
\begin{itemize} 
\item Extroversion: Strongest predictor of group formation; individuals with similar levels are more likely to connect \citep{harris_friendship_2016}. 
\item Agreeableness: Enhances group cohesion but has mixed effects on initial formation \citep{harris_friendship_2016,laakasuo_homophily_2020}. High similarity improves teamwork outcomes \citep{peeters_personality_2006}. 
\item Conscientiousness: Similarity fosters stable, high-
functioning groups \citep{laakasuo_homophily_2020,peeters_personality_2006}. However, pairing students with differing levels may aid low-Conscientiousness individuals' academic performance \citep{Poropat2009,alotaibi_benefits_2023}, though it can been observed that this carries a high risk of burdening the high-Conscientiousness peers \citep{alotaibi_benefits_2023}. 
\item Neuroticism: Some evidence suggests similarity predicts group formation \citep{laakasuo_homophily_2020}, but ethical concerns arise as Neuroticism correlates with mental health disorders \citep{jeronimus_neuroticisms_2016,kotov_linking_2010}, making its use in social recommendations problematic. 
\item Openness: No direct role in group formation, but friendships tend to form among individuals with similar levels \citep{harris_friendship_2016}. 
\end{itemize}
While all five traits contribute to social dynamics, Extroversion and Openness show the strongest evidence for homophily in group formation with the lowest ethical risk. Agreeableness, though less predictive of initial formation, supports long-term group stability and collaboration. Neuroticism and Conscientiousness pose ethical concerns, as Neuroticism correlates with mental health risks \citep{jeronimus_neuroticisms_2016,kotov_linking_2010} and Conscientiousness pairing risks counter-productive academic results for the better-performing students \citep{alotaibi_benefits_2023}.

\subsection{Theory of Mind in AI Social Agents}
Theory of Mind (ToM) refers to the ability to attribute mental states, such as beliefs, intentions, and emotions, to others and predict their behavior; In AI, ToM capabilities enable systems to model users’ cognitive and social traits for more meaningful interactions \citep{cuzzolin_knowing_2020,williams_supporting_2022}. AI-based matchmaking systems, which often lack a decent ToM, may lead to superficial matches, reinforced biases, limited adaptability, and related issues \citep{lex_psychology-informed_2021, langley_theory_2022,cuzzolin_knowing_2020}.

One recent theory in this area is the idea of a 'Mind-space', an internal representation of user traits and behaviors \citep{conway_understanding_2019}. This presented idea argues that older ToM frameworks fail to fully encompass the various person-specific traits and characteristics that influence behavior and how these are part of others' ToM to accurately predict that behavior \citep{conway_understanding_2019}. From this perspective and based on observations on general ToM and AI, systems leaning on this Mind-space idea would be able to provide deeper personalized recommendations that reflect psychological compatibility rather than surface-level preferences, and would thus more closely model real human cognition in social spaces. One facet of this deeper, personalized psychology may be inferred via the set of behaviors represented as 'personality traits'.

\subsection{Related Work}
In this work, we focus on the 44 question version of the Big-Five Personality Inventory \citep{john_bigfive_1999}, or BFI-44, to measure personality traits. This was chosen given that it is both one of the oldest and most rigorously tested scales, with proven high validity. The general Five Factor model BFI rests upon was also observed as the scale used in all of the current state-of-the-art models.

Personality inference from text has evolved from lexicon-based methods \citep{pennebaker2001liwc} to deep learning approaches \citep{mehta_bottom-up_2020}. Mehta et al. \citep{mehta_bottom-up_2020} benchmarked various models, including BERT + MLP ensembles, on the Essays dataset, showing that transformer-based models outperformed prior CNN and RN-N approaches. However, their study did not evaluate GPT-based models or personality inference on student introduction posts. Other, more recent, studies have demonstrated either comparable or slightly improved average performance over these models on various datasets. Of these, CNN-Adaboost represents perhaps the current state of the art, with an average accuracy of 61.87\% for CNN-Adaboost-2channel \citep{mohades_deilami_contextualized_2022}, though another study also showed improvement when applied to Facebook data, with XGBoost showing an average accuracy of 63-71\% depending on dictionary or network analysis technique it is combined with \citep{tadesse_personality_2018}.

To our knowledge, no prior study has investigated personality inference on student introduction posts. These posts are a common, low-effort method of student self-identification in online courses, specifically designed by the students themselves to facilitate social connection. This makes them an ideal source for unobtrusive personality assessment, while leveraging an already common characteristic of online courses. Our work fills this research gap by benchmarking both custom BERT-based and GPT-based models in this domain, applying them to student introduction post data, and evaluating their potential for AI-driven social recommendations in an educational space.

\section{Methodology}
\subsection{Dataset}
We collected 394 anonymized responses to a BFI-44 questionnaire \citep{john_bigfive_1999} from students across two semesters of Georgia Tech’s OMSCS Knowledge-Based AI course, 226 of which made introduction posts in their online course forum. For these 226 participants, their survey responses and introduction posts were combined into our dataset. Participation was voluntary, incentivized with bonus points added to their course grade, and surveys were distributed via a link on the online course forum. Students provided informed consent, explicitly agreeing to data use for research related to SAMI through survey participation. All procedures were reviewed and approved by our Institutional Review Board (IRB), ensuring compliance with ethical research standards. All 226 data points were complete and used in our study. Table~\ref{tab:trait_stats} shows the statistical description for each trait in the dataset.

\begin{table}[h]
\renewcommand{\arraystretch}{1.2}
\centering
\caption{Descriptive Statistics for Big Five Personality Traits. Trait abbreviations: O = Openness, C = Conscientiousness, E = Extroversion, A = Agreeableness, N = Neuroticism.}
\label{tab:trait_stats}
\begin{tabular}{lccccc}
\toprule
\thead{Statistic} & \thead{O} & \thead{C} & \thead{E} & \thead{A} & \thead{N} \\
\midrule
Count & 226 & 226 & 226 & 226 & 226 \\
Mean & 3.69 & 3.63 & 3.01 & 3.79 & 2.88 \\
Std Dev & 0.52 & 0.56 & 0.73 & 0.60 & 0.76 \\
Min & 2.10 & 2.00 & 1.00 & 1.90 & 1.00 \\
25\% & 3.40 & 3.20 & 2.50 & 3.30 & 2.40 \\
50\% & 3.70 & 3.70 & 3.00 & 3.80 & 2.90 \\
75\% & 4.00 & 4.00 & 3.50 & 4.20 & 3.50 \\
Max & 5.00 & 4.90 & 4.90 & 5.00 & 5.00 \\
\bottomrule
\end{tabular}
\end{table}

It was observed from this data that the traits followed distinctly different distributions. These distributional properties may have influenced model performance. Specifically, the left-skewed nature of Openness, Conscientiousness, and Agreeableness may have contributed to models more accurately predicting high values for these traits. In contrast, Extroversion, which was approximately normally distributed, posed greater challenges for classification. Neuroticism exhibited a right-skew and was the most difficult trait for models to predict accurately. We used this dataset as the ground-truth for our model tests without modification to examine model behavior on a real-world sample.

It is important to note that student introduction posts represent a distinct text type. These posts are typically written to present students favorably to their peers and contain substantial demographic information. Students commonly share their location, hobbies, career background, and interests; Some posts consist almost entirely of such demographic attributes. This distinguishes introduction posts from other text types commonly used in personality inference research, such as essays or social media content, and may influence both model behavior and predictive ability.

The dataset, implementation, and SAMI code-base used in this study are owned by Georgia Tech and are subject to institutional privacy and intellectual property restrictions. Thus, they cannot be publicly released. However, a detailed description of our methodology is provided to enable independent replication.

\subsection{Personality Detection Model}
\subsubsection{Personality trait scale}
We conducted preliminary tests to evaluate the efficacy of different classification cut-offs on a single model, GPT-4o Mini, for detecting personality traits from student introduction posts in our ground truth dataset. We examined 3 scales: a 1-5 scale, a trinary low/middle/high scale, and a binary low/high scale. For the two categorical scales, we converted categories to numeric values: 'Low' was represented by 0, 'High' by 1, and 'Middle' (for the trinary scale) by 0.5. For the low/high scale, the midpoint of the allowed trait ranges was used. We ultimately elected to use the Low/High scale as a benchmark and in system integration, due to the superior performance of the model on it. Detailed cut-offs for each trait within each scale are provided in Table~\ref{tab:compressed_range_classifications}.

\begin{table}[h]
\renewcommand{\arraystretch}{1.2}  
\caption{Trait Classifications for the 2 Performant Scales. \\ 
O = Openness, C = Conscientiousness, E = Extroversion, A = Agreeableness, N = Neuroticism}
\label{tab:compressed_range_classifications}
\centering
\renewcommand{\arraystretch}{1.3}
\setlength{\tabcolsep}{4pt} 
\begin{tabular}{l|l|l}
\hline
\thead{Trait} & \thead{Low/Med/High Scale} & \thead{Low/High Scale} \\
\hline
O & 
\begin{tabular}{l}
$0$ if $x \leq 23.33$ \\ 
$0.5$ if $23.33 < x \leq 36.67$ \\ 
$1$ if $x > 36.67$
\end{tabular} & 
\begin{tabular}{l}
$0$ if $x < 30$ \\ 
rand. $0$/$1$ if $x = 30$ \\ 
$1$ if $x > 30$
\end{tabular} \\
\hline
C & 
\begin{tabular}{l}
$0$ if $x \leq 21$ \\ 
$0.5$ if $21 < x \leq 33$ \\ 
$1$ if $x > 33$
\end{tabular} & 
\begin{tabular}{l}
$0$ if $x < 27$ \\ 
rand. $0$/$1$ if $x = 27$ \\ 
$1$ if $x > 27$
\end{tabular} \\
\hline
E & 
\begin{tabular}{l}
$0$ if $x \leq 18.67$ \\ 
$0.5$ if $18.67 < x \leq 29.33$ \\ 
$1$ if $x > 29.33$
\end{tabular} & 
\begin{tabular}{l}
$0$ if $x < 24$ \\ 
rand. $0$/$1$ if $x = 24$ \\ 
$1$ if $x > 24$
\end{tabular} \\
\hline
A & 
\begin{tabular}{l}
$0$ if $x \leq 21$ \\ 
$0.5$ if $21 < x \leq 33$ \\ 
$1$ if $x > 33$
\end{tabular} & 
\begin{tabular}{l}
$0$ if $x < 27$ \\ 
rand. $0$/$1$ if $x = 27$ \\ 
$1$ if $x > 27$
\end{tabular} \\
\hline
N & 
\begin{tabular}{l}
$0$ if $x \leq 18.67$ \\ 
$0.5$ if $18.67 < x \leq 29.33$ \\ 
$1$ if $x > 29.33$
\end{tabular} & 
\begin{tabular}{l}
$0$ if $x < 24$ \\ 
rand. $0$/$1$ if $x = 24$ \\ 
$1$ if $x > 24$
\end{tabular} \\
\hline
\end{tabular}
\end{table}

\subsubsection{Model Analysis}
While prior state-of-the-art models \citep{mehta_bottom-up_2020,tadesse_personality_2018,mohades_deilami_contextualized_2022} have demonstrated strong performance on personality inference from text, their applicability to student introduction posts in a live social recommendation system remains uncertain. Misclassification in this context could negatively impact user perception and match cohesion, particularly if personality-based recommendations do not align with user expectations.

Among these models, XGBoost + SNA \citep{tadesse_personality_2018} achieved performance comparable to our own models, suggesting that structured social network features may be highly predictive for personality inference. However, XGBoost was trained on Facebook interactions, a very different setting from our student introduction posts, raising concerns about its domain transferability. Similarly, BERT-base + MLP \citep{mehta_bottom-up_2020} and CNN-Adaboost-2channel \citep{mohades_deilami_contextualized_2022} performed well on their respective datasets. Given Mehta et al.'s results, we focused on their best-performing MLP model. 

We developed personality models leveraging various large language models (LLMs), testing them both as standalone classifiers and as base models for training. We also constructed bagging ensemble Multilayer Perceptron (MLP) models trained on LLM outputs atop LLM base models to assess whether integrating zero-shot LLM predictions with a trainable classifier improved accuracy over purely zero-shot methods, inspired by the work of Mehta et. al. \citep{mehta_bottom-up_2020}. We evaluated our models against our ground truth dataset. Table~\ref{tab:models_tested} provides a summary of these, including brief descriptions of each.

\begin{table}[h]
\setlength{\tabcolsep}{3pt}  
\renewcommand{\arraystretch}{1.0}  
\centering
\caption{Models Tested}
\label{tab:models_tested}

\begin{tabular}{ll}
    \toprule
    \thead{Model Name} & \thead{Description} \\
    \midrule
    GPT-3.5-turbo & Zero-shot GPT-3.5-turbo \\
    GPT-4o-mini  & Zero-shot GPT-4o-mini \\
    BERT & BERT (Pretrained) \\
    GPT-3.5-turbo MLP & GPT-3.5-turbo + MLP (Bagging) \\
    GPT-4o-mini MLP & GPT-4o-mini + MLP (Bagging) \\
    BERT MLP & BERT (Pretrained) + MLP (Bagging) \\
    \bottomrule
\end{tabular}
\end{table}

We benchmarked our models against the best-performing state-of-the-art (SOTA) models across multiple studies: BERT-base + MLP \citep{mehta_bottom-up_2020}, XGBoost + Social Network Analysis (SNA) \citep{tadesse_personality_2018}, and CNN-Adaboost-2channel \citep{mohades_deilami_contextualized_2022}, representing transformer-based, boosting-based, and hybrid deep learning approaches, respectively.

Ideally, direct benchmarking on a shared dataset would be preferable, but this was infeasible for two reasons:

\begin{itemize}
    \item Hardware limitations prevented large-scale retraining of Mehta et al.'s MLP models \citep{mehta_bottom-up_2020}.
    \item The dataset was found to be incompatible. Our student introductions were too small for fine-tuning and lacked the required features for the other models. XGBoost + SNA relies on social network data (user interactions, engagement), which our dataset does not contain. CNN-Adaboost-2channel extracts linguistic features from long-form text or speech transcription, making it ill-suited for brief introduction posts.
\end{itemize}

Adapting these models would require substantial computational resources and careful consideration of what additional data could be ethically collected from students. Instead, we kept these models in their original contexts to use as benchmarks while testing our models on student introductions, to ensure a fair comparison across domains. Table~\ref{tab:bert_gpt_comparison} presents the results, offering insights into the effectiveness of different modeling strategies for personality inference in real-world social recommendation systems.
\begin{table*}
\renewcommand{\arraystretch}{1.2}  
    \centering
    \begin{tabular}{lcccccc}
        \hline
        \thead{MODEL} & \thead{O} & \thead{C} & \thead{E} & \thead{A} & \thead{N} & \thead{Average} \\
        \hline
        \textbf{Trained on Essays} & & & & & & \\
        BERT-base + MLP \citep{mehta_bottom-up_2020} & 64.6 & 59.2 & 60.0 & 58.8 & 60.5 & 60.6 \\
        CNN-Adaboost-2channel \citep{mohades_deilami_contextualized_2022} & 60.56 & 64.93 & 61.85 & 59.92 & 62.08 & 61.62 \\
        \hline
        \textbf{Trained on Facebook Interactions} & & & & & & \\
        XGBoost + SNA \citep{tadesse_personality_2018} & 73.3 & 69.8 & 78.6 & 65.3 & 68.0 & 71.0 \\
        \hline
        \textbf{Trained on Student Introduction Posts (Ours)} & & & & & & \\
        BERT MLP & 86.96 & 82.61 & 43.48 & 21.74 & \textbf{56.52} & 58.26 \\
        GPT-4o Mini MLP & 86.96 & 82.61 & 58.70 & 47.83 & \textbf{56.52} & 66.52 \\
        GPT-3.5 Turbo MLP & 86.96 & 82.61 & 50.00 & 19.57 & \textbf{56.52} & 59.13 \\
        BERT & 70.35 & 25.22 & 50.44 & 65.49 & 55.31 & 53.36 \\
        GPT-4o Mini & 90.71 & \textbf{84.07} & \textbf{60.18} & \textbf{90.27} & 53.10 & \textbf{75.66} \\
        \textbf{GPT-3.5 Turbo} & \textbf{91.59} & \textbf{84.07} & 57.96 & \textbf{90.27} & 52.65 & 75.31 \\
        \hline
    \end{tabular}
    \caption{Comparison of Accuracies of Models on Different Datasets. Trait Abbreviations: O = Openness, C = Conscientiousness, E = Extroversion, A = Agreeableness, N = Neuroticism.}
    \label{tab:bert_gpt_comparison}
\end{table*}

We did not modify the introduction post text in the ground-truth data in any way, for any of the models tested. The introduction posts were passed into the models as-is, and thus did not have any stop-words removed, or any structure modified as might have been usual for natural language processing. This decision was based on prior research observing that word choice, sentence structure, and linguistic patterns contribute to personality inference \citep{mairesse_using_2007}. Modifying the raw text may thus compromise important signals that models use to predict personality traits.
 
\begin{algorithm}[h]
\renewcommand{\arraystretch}{1.2}  
\caption{Personality Extraction Using GPT}
\label{alg:personality-extraction}
\textbf{Require:} Student introduction post as string \textit{text}, API key \textit{api\_key} \\
\textbf{Ensure:} Big-5 personality traits (\textit{Openness, Conscientiousness, Extroversion, Agreeableness, Neuroticism}) with values ``low'' or ``high''
\begin{algorithmic}[1]
    \STATE Define traits: $\mathcal{P} = \{$Openness, Conscientiousness, Extroversion, Agreeableness, Neuroticism$\}$
    \STATE Define output format: $\{ p : v \mid p \in \mathcal{P}, v \in \{\text{``low'', ``high''}\} \}$
    \STATE Set system role description: ``You are an expert in inferring students' Big-5 personality traits from text that they have written.''
    \STATE Construct user role query: ``For the given text sample, infer the author's personality traits and return your results in the following format without explanation: '' + \textit{text}
    \STATE Send query to the language model via the OpenAI API, using the following parameters:
\begin{compactitem}
    \item Model: \texttt{model\_name}
    \item Temperature: 0
    \item Prompts: System and user prompts, as defined in 3 and 4 above
\end{compactitem}
    \STATE Parse \textit{output} $\leftarrow$ API response
    \STATE \textbf{return} \textit{output}
\end{algorithmic}
\end{algorithm}
BERT-Base-Personality\footnote{The BERT-Base-Personality model is hosted on Hugging Face and was accessed on August 24th, 2024. URL: \url{https://huggingface.co/Minej/bert-base-personality}.}, which is pretrained to predict the Big Five personality traits \citep{devlin_bert_2019}. We made no modifications to this base model. For the bagging ensemble models (MLP), we first used the LLM model to predict a binary High/Low score for each personality trait, which was then treated as the input feature vector (\textit{X}) for the bagging ensemble models. A separate bagging ensemble Multilayer Perceptron (MLP) model was trained and tested for each of the Big Five personality traits. Both \textit{X} and the target labels (\textit{y}) were split into training and testing sets using an 80/20 train-test split of the ground-truth dataset.

We did not train the base GPT models, as they demonstrated high efficacy in zero-shot personality trait inference. 

In LLMs like GPT, zero-shot learning refers to the model's ability to generalize to new tasks without fine-tuning or task-specific training. Consequently, we did not split the ground-truth data into training and testing sets for these models. Instead, we directly compared their predictions to actual personality traits across the full dataset. Unlike our MLP models, which required supervised learning with a train-test split, the base-only GPT models were evaluated as-is without additional training. All GPT models used the same system prompt: "You are an expert in inferring students' Big-5 personality traits from text that they have written." The complete prompt, and all hyper-parameters, are displayed in Algorithm~\ref{alg:personality-extraction}, above.

\subsection{SAMI: Prototype for Incorporating Personality Traits}
We integrated our chosen personality detection model within SAMI's entity detection component. This component extracts entities from students' introduction posts, which are then stored as nodes in a graph database \citep{kakar2024sami}. Our model runs after this step and personality traits are appended as entities to the entities for a student. Due to this, SAMI treats a student's personality trait data like other entities, such as hobbies, performing homophily matching using personality traits along with all other entities, as outlined in SAMI's introductory paper \citep{kakar2024sami}. We selected 3 of the 5 personality traits detected to be saved and passed on to the matchmaking, due to their relevance noted in the Background section above to similarity (homophily) matching and group cohesion: Extroversion, Agreeableness, and Openness.

Within the current matchmaking system in SAMI, each of the 3 personality traits selected is treated as a distinct entity. The weight assigned to these traits, which is determined when SAMI selects the top 5 matches, is based on the prevalence of the trait level in the current student body interacting with SAMI in a course. SAMI’s selection process ranks potential matches based on a composite score that considers entities such as hobbies and location, as detailed in \citep{kakar2024sami}. Our proposed system matches on personality traits alongside the existing metrics. 

We hypothesized that personality, alongside hobby alignment, is one of the most powerful indicators of successful social matches. We further hypothesize that treating these personality traits as such in the matchmaking weighting will result in recommendations more closely modeling organic social matches. To implement this, we assigned the relative weight of all 3 traits together to be the 2nd most important entity in this process, when these and the original entities are included. Depending on the relative rarity of hobbies and/or personality trait levels to the overall class distribution, this weight varies and the combined weights of personality traits may be the 1st or 3rd most important in the matchmaking algorithm for some cohorts.

\begin{table}[h]
\centering
\caption{Synonym List for Personality Traits. 
E = Extroversion, A = Agreeableness, O = Openness.}
\label{tab:synonyms}
\begin{tabular}{|l|l|p{5cm}|} 
\hline
\thead{Trait} & \thead{Level} & \thead{Synonyms} \\ 
\hline
\multirow{2}{*}{\textbf{E}}  
    & High & sociable, outgoing, gregarious, charismatic, lively, expressive, energetic, enthusiastic, talkative, friendly 
    \\ \cline{2-3} 
    & Low  & reserved, quiet, observant, introspective, thoughtful, calm, reflective, private, contemplative, introverted, low-key 
    \\ 
\hline
\multirow{2}{*}{\textbf{A}}  
    & High & kind, cooperative, empathetic, warm, compassionate, friendly, generous, understanding, supportive, helpful  
    \\ \cline{2-3} 
    & Low  & independent, confident, self-reliant, forthright, direct, strong-willed, principled, uncompromising, determined, self-assured 
    \\ 
\hline
\multirow{2}{*}{\textbf{O}}  
    & High & imaginative, inventive, curious, innovative, inquisitive, adventurous, visionary, creative, unconventional, explorative  
    \\ \cline{2-3} 
    & Low  & pragmatic, consistent, stable, familiar, traditional, secure, steady, reliable, grounded, predictable 
    \\ 
\hline
\end{tabular}
\end{table}

Additionally, to ensure transparency, students are presented with their detected personality traits, as well as which they share with their presented matches. These were implemented as simple additions to SAMI's original summary and match feedback response features. To mitigate potential negative connotations (e.g., "low agreeableness") and cultural biases, raw trait descriptions are replaced with synonyms, randomly selected from a list of predefined dictionaries. Thus, users will not see the raw traits and levels, only the favorable synonyms. The full contents of the dictionaries containing all synonyms from which a replacement is selected may be seen in Table~\ref{tab:synonyms}, above.

This prototype provides a valuable proof-of-concept for how our personality model and method of collecting student personality trait data may be used in recommendation system deployed on a live online class setting. For the integration, we selected traits with established connections to positive social formation, recognizing that deployment in live educational settings will require validation of both prediction biases and student outcomes.

\section{Results and Discussion}
\subsection{Personality Model}
As noted in Methodology, we performed tests to select the best scale to use with our models. The results of these tests are presented in Table~\ref{tab:accuracy_f1_scores}, though the 1-5 scale results were excluded due to extremely low (approaching 0) f1 scores across all traits. The binary high/low scale demonstrated the highest performance, with the other scales yielding significantly lower results. We suspect that alignment with the traits’ distribution of our dataset was the likely cause.
\begin{table}[h]
\setlength{\tabcolsep}{3pt}  
\renewcommand{\arraystretch}{1.1}  
\centering
\caption{Accuracy and F1 Scores for Two Scales Using Zero-shot GPT-4o Mini. Traits: O (Openness), C (Conscientiousness), E (Extroversion), A (Agreeableness), N (Neuroticism).}
\label{tab:accuracy_f1_scores}

\begin{tabular}{lcc|lcc}
    \toprule
    \multicolumn{3}{c|}{\textbf{Low | High Scale}} & \multicolumn{3}{c}{\textbf{Low | Middle | High Scale}} \\
    \midrule
    \thead{Trait} & \thead{Acc} & \thead{F1} & \thead{Trait} & \thead{Acc} & \thead{F1} \\
    \midrule
    O  & 0.91 & 0.95  & O  & 0.55 & 0.63  \\
    C  & 0.84 & 0.91  & C  & 0.58 & 0.53  \\
    E  & 0.60 & 0.66  & E  & 0.56 & 0.26  \\
    A  & 0.90 & 0.95  & A  & 0.54 & 0.58  \\
    N  & 0.53 & 0.02  & N  & 0.28 & 0.00  \\
    \midrule
    \textbf{Overall} & 0.76 & 0.70  & \textbf{Overall} & 0.50 & 0.40  \\
    \bottomrule
\end{tabular}
\end{table}

Table~\ref{tab:model_performance} shows the variation between traits across all models. While both the BERT and MLP models demonstrate strong performance in this task, the base GPT models outperformed them, with the GPT-4o-mini model achieving the highest accuracy and F1 scores.

\begin{table*}
\renewcommand{\arraystretch}{1.2}
\centering
\caption{Model Performance by Trait (Accuracy and F1 Score). 
Traits: O = Openness, E = Extroversion, A = Agreeableness, C = Conscientiousness, N = Neuroticism.}
\label{tab:model_performance}

\begin{tabular}{lrrrrrrr}
\toprule
\thead{Model} & \rotatebox{0}{O (Acc/F1)} & \rotatebox{0}{E (Acc/F1)} & \rotatebox{0}{A (Acc/F1)} & \rotatebox{0}{C (Acc/F1)} & \rotatebox{0}{N (Acc/F1)} & \rotatebox{0}{Average} \\
\midrule
BERT MLP          & 86.96/93.02 & 43.48/31.58 & 21.74/25.00 & 82.61/90.48 & 56.52/72.22 & 58.26/62.46 \\
GPT-4o Mini MLP   & 86.96/93.02 & 58.70/66.67 & 47.83/55.56 & 82.61/90.48 & 56.52/72.22 & 66.52/\textbf{75.56} \\
GPT-3.5 Turbo MLP & 86.96/93.02 & 50.00/53.06 & 19.57/5.13  & 82.61/90.48 & 56.52/72.22 & 59.13/67.78 \\
BERT              & 70.35/82.32 & 50.44/60.00 & 65.49/78.45 & 25.22/28.69 & 55.31/62.17 & 53.36/62.33 \\
GPT-4o Mini       & 90.71/95.13 & 60.18/66.42 & 90.27/94.86 & 84.07/91.30 & 53.10/1.85  & \textbf{75.66}/69.91 \\
GPT-3.5 Turbo     & 91.59/95.61 & 57.96/54.55 & 90.27/94.88 & 84.07/91.35 & 52.65/0.00  & 75.31/67.28 \\
\bottomrule
\end{tabular}

\end{table*}

Reviewing the trait breakdown, model performance varied across traits. While GPT models demonstrate high accuracy for certain traits, several factors warrant careful interpretation. We observe that model performance varied across traits based on their distributions. Our models achieved high accuracy on left-skewed traits (Openness, Agreeableness, Conscientiousness), but more modest performance on normally-distributed Extroversion, and no real predictive capability for right-skewed Neuroticism. On the surface, this would seem to indicate better than SOTA performance, but consider that a naive classifier predicting only the majority class would be expected to achieve 75-80\% accuracy on O, C, and A. This indicates that while the models exceed this as much as 11\%-16\%, the margin is much smaller than accuracy alone would suggest. This points to both a strong 'optimism' bias and genuine feature extraction. As noted previously, student introduction posts are inherently demographic-heavy, which may also influence model predictions. Importantly, performance on the most normally-distributed trait, Extroversion, is comparable to SOTA models with a good F1.

Despite the distributional issues, we selected the GPT-4o-mini base model for the SAMI matchmaking system above. This decision was based on its superior performance on Extroversion, a trait that is also perhaps the most important for social cohesion as noted earlier. While we observed a high likelihood of majority-class bias in the left-skewed traits, GPT-4o-mini demonstrated the most consistent predictive ability across traits.

\subsection{Ethics}
AI social recommendation systems like SAMI must address ethical challenges when using personality traits for matchmaking, including privacy, inappropriate recommendations, user autonomy, and ethical data use \citep{baum_social_2020, dignum_responsible_2017, vetro_ai_2019}.

Relying solely on personality traits risks stereotyping and altering student social dynamics. To counter this, our modified SAMI incorporates hobbies and other attributes for balanced recommendations. It focuses on Extroversion, Agreeableness, and Openness, omitting Conscientiousness due to potential academic performance bias and Neuroticism due to ethical concerns related to mental health.

Regarding privacy, SAMI initially shared only voluntarily disclosed data. However, inferred personality traits may reveal sensitive characteristics \citep{tkalcic_personality_2015}. To mitigate cultural biases, SAMI presents only favorable synonyms, though research on its impact is limited. The main privacy risk lies in using GPT-4o-mini for personality inference, requiring third-party processing. However, no PII is shared, and names are anonymized before sending. While this trade-off currently balances performance and privacy, further mitigation strategies should be discussed.

\section{Limitations and Future Work}
A limitation of this study is the absence of real-world performance testing to evaluate the impact of incorporating personality data into SAMI's matchmaking system. Additionally, we currently lack the ability to assess the system's fairness across different demographics due to the limited sample size available within our research group. The long-term effects of personality-informed matchmaking on social group dynamics also remain untested.

Our best-performing model demonstrates weaker predictive capability for Extroversion compared to the other two selected traits. Moreover, model performance varies systematically with trait distributions. The model's strong performance on certain traits may stem from distribution bias, raising concerns about its generalizability beyond this specific course. These considerations motivate ongoing work examining class-specific performance metrics and the reasoning patterns underlying model predictions.

To address these limitations, several directions for future work emerge. First, to improve Extroversion prediction to exceed SOTA, we are exploring prompt engineering techniques and investigating hybrid models. Second, future work should analyze trait distributions across broader populations, institutions, and multiple courses to determine whether the observed skewness reflects real-world trends or is dataset-specific. Third, controlled deployment studies examining student engagement, perceived usefulness of recommendations, and social connection outcomes are needed to validate the proof-of-concept architecture when comparing the personality-enhanced system against the baseline. Finally, self-hosted LLM models may be viable alternatives for personality inference. These models could provide a more privacy-conscious approach while reducing cost, though validation of their predictive performance on this task is necessary.

\section{Conclusion}
In this work, we explored the integration of personality detection within SAMI. By incorporating personality-enhanced recommendations, we aimed to improve SAMI's matchmaking capabilities by more closely emulating organic social selection in the wild. We focused on traits with proven strong correlations to beneficial social interaction and to grouping behavior while avoiding those that could pose ethical concerns.

We demonstrate that zero-shot GPT models can perform personality inference unobtrusively on introduction posts, achieving performance results comparable to established benchmarks from other text domains. In particular, we found that Extroversion could be predicted with accuracy comparable to SOTA models on the novel text format of organic student introduction posts. This work establishes the technical feasibility of incorporating personality inference into social recommendation systems, though both validation of real-world effectiveness and deep analysis of model interpretability remain future work. While limitations remain, our work provides a foundation for future research into AI-driven matchmaking and personality inference.

\renewcommand{\bibsection}{\section{REFERENCES}}
\bibliographystyle{abbrv}
\bibliography{personality-enhanced-social-rec}

\end{document}